\pdfoutput=1
\documentclass[11pt]{article}

\usepackage{acl} %[review]

\usepackage{times}
\usepackage{latexsym}
\usepackage{annotates}
\usepackage{amsfonts}

\usepackage[T1]{fontenc}
\usepackage[utf8]{inputenc}

\usepackage{microtype}

\usepackage{inconsolata}
\usepackage{hyperref}
\usepackage{todonotes}
\usepackage{multirow}
\usepackage{booktabs}
\usepackage{xspace}
\usepackage{amsmath}

\newcommand{\SV}{SBERT\textsubscript{Vanilla}\xspace}
\newcommand{\SD}{SBERT\textsubscript{Domain}\xspace}
\newcommand{\SH}{SBERT\textsubscript{Hashtag}\xspace}
\newcommand{\SP}{SBERT\textsubscript{Party}\xspace}

\title{Toeing the Party Line: Election Manifestos as a Key to Understand Political Discourse on Twitter}

\author{
Maximilian Maurer\textsuperscript{$\triangle$}, Tanise Ceron\textsuperscript{$\ddagger$}, Sebastian Pad\'o \textsuperscript{$\square$} \normalfont{and} \textbf{Gabriella Lapesa\textsuperscript{$\triangle$}}\\
\textsuperscript{$\triangle$}GESIS and Heinrich-Heine University Düsseldorf, Germany\\
\textsuperscript{$\ddagger$}Bocconi University, Italy\\
\textsuperscript{$\square$}Institute for Natural Language Processing, University of Stuttgart, Germany\\
\texttt{first.last@gesis.org}, \texttt{tanise.ceron@unibocconi.it}, \texttt{pado@ims.uni-stuttgart.de}
}

\begin{document}

\maketitle

\begin{abstract}
Political discourse on Twitter is a moving target: politicians continuously make statements about their positions. It is therefore crucial to track their discourse on social media to understand their ideological positions and goals.
However, Twitter data is also challenging to work with since it is ambiguous and often dependent on social context, and consequently, recent work
on political positioning has tended to focus strongly on manifestos (parties' electoral programs) rather than social media. \\
In this paper, we extend recently proposed methods to predict pairwise positional similarities between parties from the manifesto case to the Twitter case, using hashtags as a signal to fine-tune text representations, without the need for manual annotation.
We verify the efficacy of fine-tuning and conduct a series of experiments that assess the robustness of our method for low-resource scenarios.
We find that our method yields stable positioning reflective of manifesto positioning, both in scenarios with all tweets of candidates across years available and when only smaller subsets from shorter time periods are available. This indicates that it is possible to reliably analyze the relative positioning of actors forgoing manual annotation, even in the noisier context of social media.
\end{abstract}

\section{Introduction}
In the domain of political science, manifestos have been commonly utilized as a means to analyze the positioning of political parties. This type of document lends itself well to the task given that it is an official, stable, and compromised textual source with clearly stated stances, ideologies and preferences \citep{bakker-2013,DiCocco_Monechi_2022,ademi-24}. However, while understanding the positioning of political actors through manifestos remains relevant, there are also some pitfalls in reducing the analysis to this type of genre only. First, manifestos are only published in proximity to elections (e.g., every 4 years for German federal elections), while tweets allow for continuous, fine-grained temporal analysis); second, tweets may uncover issues beyond the official position of a party as agreed on and curated in the form of manifestos \citep{birdfish}; third, and crucially, large parts of the electorate do not read manifestos released from parties, as polls suggest,\footnote{For instance, two-thirds of the British public either do not read party manifestos or do not know what they are. \url{https://www.bmgresearch.co.uk/bmg-research-poll-10-people-dont-know-manifesto/}} making manifestos a weak model of the public political discourse consumption.

\begin{figure}[b!t]
    \centering
    \includegraphics[width=\linewidth]{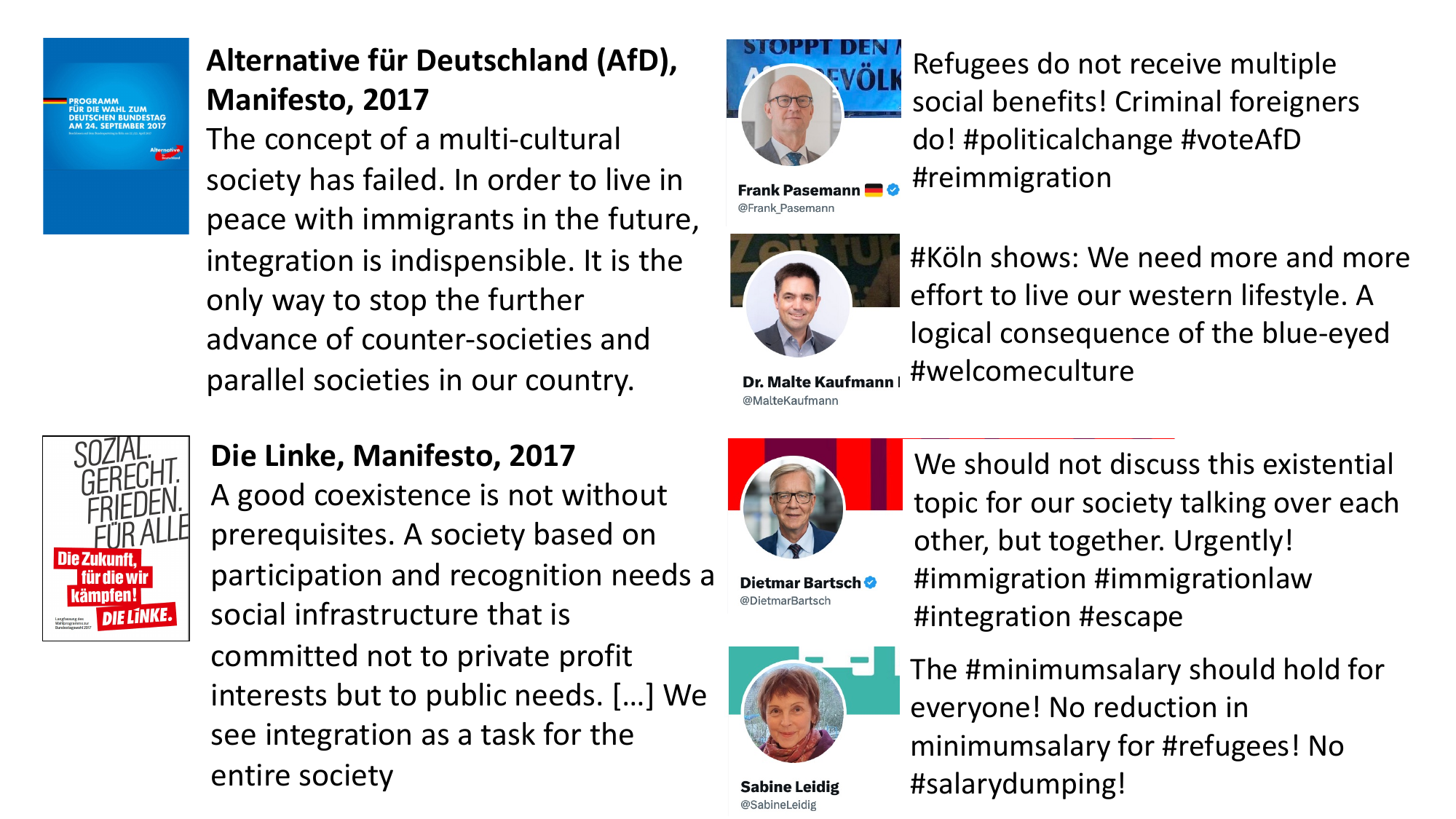}
    \caption{Political discourse: Manifestos vs. Twitter}
    \label{fig:examplemanifesto}
\end{figure}

In the digital age, short-text social media platforms such as X (formerly Twitter) have become an increasingly important communication channel for political actors (politicians, parties, institutions) to convey their stances on policy issues to potential voters (e.g., raising the minimum wage; closing borders to refugees).
Besides that, the public increasingly relies on social media for political information and opinion formation \citep{weeks2017online}. 
For those reasons, automating the analysis of online political debates adds to the positioning analysis, providing a clearer understanding of the actors' stances. This is not only relevant to political scientists who monitor and investigate the political spectrum within a country, but also to citizens who then take decisions at the election time.

Having said that, analyzing social media data such as Tweets poses new challenges for computational models.
Consider Figure \ref{fig:examplemanifesto} (consult Appendix \ref{appendix:fig1} for sources), the left-hand column shows excerpts from the party manifestos for the far-right \textit{Alternative für Deutschland} party (AfD) (top) and the left-wing party \textit{Die Linke}. To the right, we show four (translated) tweets by politicians in the same party. 
Both text genres present stances towards the policy issue of migration (which are opposing views between the two parties illustrated in the example). However, the extracts from the manifestos are unambiguous and present a clear stance towards the issue, whereas the tweets leave room for interpretation which is highly dependent on social context where the meaning often goes beyond words, as shown in previous studies \citep{pujari-goldwasser-2021-understanding,hovy-yang-2021-importance}. For example, the second tweet requires knowledge about \textit{blue-eyed (i.e., naïve) welcome culture} to fully capture the positioning of the politician. Moreover, claims can be vague, such as the call for action from the third tweet.

Given that a large body of work on the automatic extraction of policy positions from texts
focuses on manifestos or legislative speeches \citep{wordfish, wordshoal, glavas-etal-2017-unsupervised, ceron-etal-2022-optimizing}, 
and in contrast, overwhelmingly avoids social media text -- a huge research area in their own right -- 
we build on previous approaches to bridge this gap to extracting the positioning of parties from tweets.  
We compute sentence embeddings for tweets, leveraging the insight that politicians may use hashtags as shorthands for policy issue areas \citep{shapiro-hemphill2017} -- as also shown in Fig. \ref{fig:examplemanifesto}. 

We carry out a series of experiments based on a large corpus of tweets from German politicians and evaluate by correlating 
party distances extracted from manually annotated manifestos \citep{manifesto}  to party distances computed from tweets of party members for the election years 2017 and 2021. 
Our experiments establish that fine-tuning with hashtags outperforms other types of fine-tuning.
We complement this quantitative result with a qualitative analysis of the hashtags in our corpus. 

We then proceed to a set of experiments in which we explore the robustness of our method to low-resource scenarios, addressing the research question of \textit{how many tweets are needed before performance degrades}. We address this question in three follow-up experiments: a) randomly sampling tweets, b) sampling tweets for time intervals farther from and closer to the election, and c) by sampling tweets for specific groups of politicians.

Our contributions are twofold: a) at the level of NLP methods, we contribute to the landscape of methods that aim at bridging between the short-term, small-size representation on Twitter and domains where longer and more curated texts are available. To the best of our knowledge, our study is the first to extract the positioning of parties through tweets in an unsupervised fashion; b) at the interdisciplinary level, we design and robustly evaluate a workflow that enables political scientists to perform theory-driven, large-scale analyses of the dynamics within the political landscape on Twitter. 

We make our code, models, and data in the form of TweetIDs and the metadata necessary for replication freely available\footnote{\href{https://github.com/mmmaurer/toeing-the-party-line}{https://github.com/mmmaurer/toeing-the-party-line}}.

\section{Related Work}

\paragraph{Automatic Political Text Scaling/Positioning.}
To reduce the need for large-scale annotation, several methods of automatic scaling and positioning of political actors have been proposed. The first automatic approaches to extract policy positions from text relied only on word count information \citep{wordscores,wordfish,wordshoal}, which are not applicable in cross-lingual settings and do not capture semantic relationships between the words in the texts.  
Then, \citet{glavas-etal-2017-unsupervised} and \citet{scaling-semantics} improve the performance of the task with word embedding-based scaling method combined with graph-based score propagation.
\citet{rheault_cochrane_2020} propose to use party embeddings retrieved from word embeddings augmented with political metadata trained on parliamentary corpora and PCA to retrieve party positions. 
Building on the advances of language models, \citet{ceron-etal-2022-optimizing,ceron-etal-2023-additive} 
experiment with fine-tuning the sentence embeddings with in-domain data from manifestos in setups with little or no annotated data. Results show a high correlation with the positioning of parties. 

\begin{figure}[t!b]
    \centering
\includegraphics[width=\linewidth]{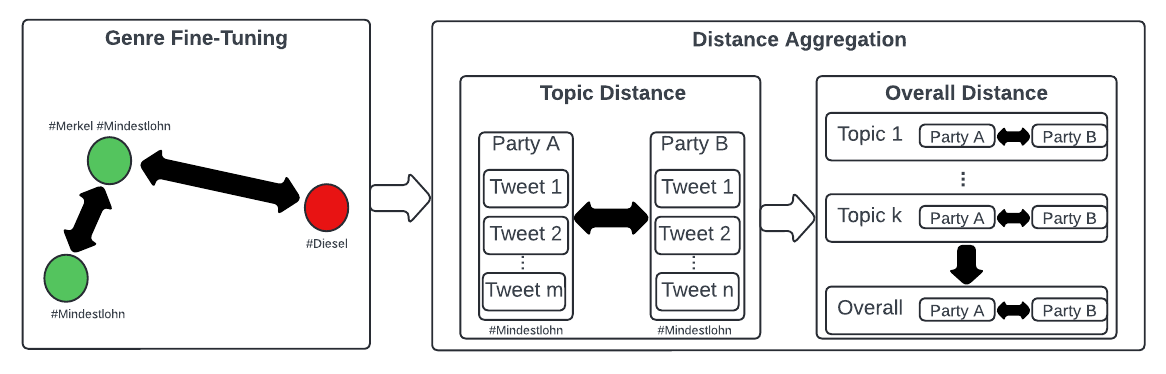}
    \caption{Illustration of our workflow.}
    \label{fig:workflow}
\end{figure}

\paragraph{Analyzing Politician's Social Media Posts.}
 \citet{socialmedia2014elections} analyze Twitter posts from political figures in the 2014 European elections in terms of hashtag usage and sentiment in their posts. They find that the content in terms of hashtags and the emotional tone of the communication of candidates is more aligned with European issues than left-to-right positions. This indicates that social media postings contain information on policy, contrary to what prior work suggests \citep{jungherr2014twitter}. Similarly, \citet{intraparty}, \citet{birdfish}, and \citet{saltzer2022bundestagswahl} use Word\-fish/Word-shoal to assess inter- and intra-party positions for social media postings of politicians before elections. The former two validate that, in tendency, social media postings reproduce inter and intra-party positions.

While most studies on social media analyze tweets from the public, a few investigate the discourse of politicians, such as analyzing hashtag and work usages \citep{green-larasati-2018-first}, detecting stances \citep{johnson-goldwasser-2016-identifying}, classifying whether individual tweets are ideological \citep{djemili-2014-twitter}, and classifying the sentiment \citep{stranisci-etal-2016-annotating} of politicians' tweets. However, in contrast with our work, they do not address the task of identifying the positioning of political actors with unsupervised or weakly-supervised methods. 

\section{Data}

\paragraph{Twitter Corpus}

We conduct our study on a corpus containing tweets by candidates of three German federal elections (2013, 2017, 2021). 
This corpus is a pooled version of three collection periods for election 2013\footnote{\href{https://doi.org/10.4232/1.12319}{https://doi.org/10.4232/1.12319}}, 2017\footnote{\href{https://doi.org/10.4232/1.12992}{hhttps://doi.org/10.4232/1.12992}}, and 2021\footnote{\href{https://doi.org/10.4232/1.14233}{https://doi.org/10.4232/1.14233}}, and in-between election Tweets collected by \citet{birdfish}, starting from 2008.
The linked DOIs have published Tweet IDs.

In total, the corpus contains over 10M Tweets. For data sparsity reasons in the earlier years, we only use data from the years 2013--2021 in this work overall, using data from the years 2017 and 2021 for the evaluation step. Table \ref{tab:dataset} in the Appendix lists the numbers of Tweets and politicians per year. We make the Tweet IDs and meta-information necessary for replicating our experiments publicly available.

\paragraph{Comparative Manifesto Project corpus}
The Comparative Manifesto Project (CMP) is one of the largest and most well-established datasets in the subfield of party competition in political science \citep{manifesto} that has evolved across countries and continents\footnote{\url{https://manifestoproject.wzb.eu/}}. It consists of manifestos extensively annotated by experts at a (quasi-)sentence level. The annotations are based on a hierarchical codebook that categorizes manifesto sentences at two different levels of granularity. The higher level covers seven broader policy domains such as \textit{external relations} or \textit{economy}. The finer-grained level encompasses 143 sub-categories within the policy domains. The annotation covers stances as well, e.g., \textit{Military:Positive} indicates that the sentence supports policies favoring the military whereas \textit{Military:Negative} refers to sentences critical of military force. In this paper, we work with the German manifestos from the elections years 2017 and 2021 for the six main German parties (CDU/CSU, SPD, FDP, Greens, Left, AfD).

\section{Method}
Our proposed method aims to gather meaningful representations and, based on those, ideological positionings of parties for the computational analysis of the political Twitter domain. Following insights on how policy is communicated on Twitter \citep{shapiro-hemphill2017}, we propose a two-step workflow.
As Figure \ref{fig:workflow} illustrates, we first gather text genre-specific sentence embeddings, as detailed
in Section \ref{sec:hashtag-tuning}. In the second step, we first gather topic-specific distances, which are then aggregated to gather overall party distances, as described in Section \ref{sec:topic-aggregation}.

\subsection{Optimizing Sentence Encoders for Genres}
\label{sec:hashtag-tuning}

We use the term \textit{genres} to denote the variability in text types, given that \textit{domains} pertain to policy-related issues and remain constant across Tweets and manifestos. As \citet{ceron-etal-2022-optimizing} show for manifesto texts, SBERT \citep{reimers-gurevych-2019-sentence} models trained on a rather general task, in principle, can be fine-tuned for the task of political positioning. Following this, we propose an SBERT fine-tuning procedure for political Twitter data, without the need for manual annotation, based on the political science insight that politicians may use hashtags as a shorthand for policy-related topics \citep{shapiro-hemphill2017}:
A pair of tweets share a topic, and their representations thus ought to be close to each other in the representation space if the tweets have at least one overlapping hashtag. In contrast, if a pair of tweets does not have co-occurring hashtags, we assume them to have no topical overlap. Thus, their representations should be further apart in the representation space. 

A qualitative analysis of the most frequent hashtags in the period spanning from 2017 to 2021 confirms the assumption by \citet{shapiro-hemphill2017} and provides further insight into the properties of our method. As shown in Table \ref{tab:example_hashtags}, besides the obvious mentions of political actors and policy domains, whose tweets equip the fine-tuned models with a party-specific and domain-specific signal which we can assume to hold for larger time spans, mentions of events such as scandals or episodes of violence is a property of a specific subset of the data in which a meaning shift can be observed, related to the hashtag ("Hanau" not denoting the city anymore, but the reference to a symbolic event). Lastly, slogan hashtags (compared to the mentions of parties) equip the representations with a finer-grained signal regarding the stance of the tweet author (as opposed to the generalized party membership encoded by the more general hashtags).     

\begin{table}[]
%\small
\begin{tabular}{p{0.18\columnwidth}p{0.7\columnwidth}}
\hline
\textbf{Category} & \textbf{Examples} \\\hline
political actors & \#SPD, \#AFD (parties) \#Merkel (politicians), \#Bundestag (institutions) \\ \hline
policy domains & \#Impfung (vaccination),  \#Mindestlohn (minimum salary), \#Seenotrettung (sea rescue), \#Kohle (coal)   \\ \hline
election-specific & \#btw17 (labels for the 2017 elections), \#GroKO, \#Jamaica, \#Ampel (possible coalition configurations after the elections), \#Triell (tv-duel among the top 3 candidates)  \\  \hline
slogans & \#nonazis, \#SPDerneuern (renew the SPD),  \#wegmit219a (referring to the abortion debate) \\  \hline
events & \#Brexit, \#Hanau (shooting by a far-right extremist in the city of Hanau, 2020), \#Dieselgate (Volks\-wagen emissions scandal, 2015) \\ \hline
\end{tabular}
    \caption{Hashtag categories and  examples}
    \label{tab:example_hashtags}
\end{table}

\subsection{Distance Aggregation}
\label{sec:topic-aggregation}
Since hashtags can be assumed to approximate topics \citep{shapiro-hemphill2017}, we hypothesize that for a pair of parties, we can 
aggregate distance among the parties' tweets topically related by a shared hashtag and then average these distances to obtain an informed approximation of how much the two parties differ in their overall political position. We compute this distance as follows:

\paragraph{1. Distance per Topic.}
Given the set of tweets with a given hashtag from party $a$, $C_a$, and the set of tweets with the same hashtag from party $b$, $C_b$, we calculate the average cosine distance between all $N$ possible  tweet pairs:

\begin{equation}
    dist(C_a, C_b)=\frac{1}{N} \sum_{s_i\in C_a, s_j \in C_b} 1-\cos(s_i, s_j)
    \label{eq:dist}
\end{equation}

\noindent
$s_i\in C_a$ and $s_j\in C_b$ refer to sentence embeddings of tweets from parties $a$ and $b$, respectively.

\paragraph{2. Overall Distance.}
To obtain overall distances between two given parties $a$ and $b$, we aggregate their distances by averaging per hashtag:
\begin{equation}
    dist(a, b) = \frac{1}{|H|} \sum_{h\in H} dist(C_{h,a}, C_{h,a})
\end{equation}

\noindent
where $h$ is one hashtag from the set of hashtags $H$. The result is $dist(C_{h,a}, C_{h,a})$, the per-topic distance between the two parties for hashtag $h$.

\section{Evaluation}
While we expect positions on Twitter to be more rich and variable on an individual politician level, party members can be expected to unite behind the \textit{party line}, especially in election campaigns, as candidates have been found to consider manifestos to be important documents for their campaigning \citep{eder-2017-manifesto}. We thus turn to manifestos as a ground truth for evaluation.

\paragraph{Ground Truth: The CMP Distance Matrix}
We use the labels annotated in the CMP to capture the parties' positioning within manifestos.  
Our ground truth is based on all fine-grained categories from the CMP codebook because it provides a full picture of the political preferences of each party rather than picking some policy issues that belong to the left or right, for example. It is also more in line with our task because we are measuring the positioning of parties across multiple policy issues.

We calculate the positioning of the parties based on the salience of the annotated fine-grained categories from the codebook of the CMP corpus. Each party is represented by a vector that considers the number of counts for each of the 143 categories normalized by the length of the manifestos. A distance matrix ($\mathbb{R}^{NxN}$), as illustrated in Figure \ref{fig:cmp-matrix} in the Appendix, is computed based on the Euclidean distance between each pair of the $N$ party vectors.

\paragraph{Evaluation Metric}
Building on previous work \citep{ceron-etal-2022-optimizing,ceron-etal-2023-additive}, to evaluate the resulting distance matrices, we use the Pearson-normalized Mantel test \citep{Mantel1967TheDO}. This type of metric allows us to correlate distance matrices given that the observations are not independent of each other. 

\section{Experiment 1:  Comparing Representation Models}
In this first experiment, we assess whether distances between parties reflect manifesto distances. Moreover, we compare sentence encoder models trained on different levels of genre information in our distance aggregation workflow to assess the impact of genre-specific information in the representations (political tweets vs. manifestos). Specifically, we compare our political Twitter-specific model to more general models and models fine-tuned in the political domain, but with a distinct type of text genre.
We assess the impact of our distance definition by comparing its correlation
with manifesto distances to a less informed aggregation of distances. 

\subsection{Experimental Setup}

\subsubsection{Evaluation Data}
We evaluate our method on data from the election years 2017 and 2021, both on the manifesto and the Tweet side. To ensure verified party membership in a given election year, we restrict the tweets considered for evaluation to the ones of the candidates in the respective year.
Given that our distance aggregation method requires hashtags to be present across all parties, we only sample those Tweets with at least one hashtag fulfilling this criterion. This results in 442k tweets being considered for 2017, and 795k for 2021.

\subsubsection{Sentence Encoder Models}

To assess the level of domain-specific information encoded in the representations necessary for meaningful ideological positioning from tweets, we experiment with three sentence-transformer models:

\noindent
\textbf{\SV.} We choose a multilingual SBERT model pre-trained on a paraphrase detection task\footnote{\href{https://huggingface.co/sentence-transformers/paraphrase-multilingual-mpnet-base-v2}{https://huggingface.co/sentence-transformers/paraphrase-multilingual-mpnet-base-v2}} as an off-the-shelf SBERT baseline. This allows us to compare a generally available standard sentence encoding model trained on data that is neither political nor from social media to models trained specifically on and for political and social media data.

\noindent
\textbf{\SD.}\footnote{\href{https://huggingface.co/tceron/sentence-transformers-party-similarity-by-domain}{https://huggingface.co/tceron/sentence-transformers-party-similarity-by-domain}} This model is based on \SV{}. To capture distances between parties, this model is fine-tuned on CMP quasi-sentences such that the embeddings of two sentences are highly similar if they have the same policy domain annotation, and have a low similarity if they do not \citep{ceron-etal-2022-optimizing}. For an overview of the policy categories and example annotations of quasi-sentences, see Appendix \ref{appendix:cmp-annot}. 

In contrast to \SV, this model encodes information about the political domain useful for policy-informed aggregation of interparty distances \citep{ceron-etal-2023-additive}, but not specifically on social media. It thus serves as a baseline for how well models trained on a different genre of political text transfer to our target genre, namely Twitter.

\noindent
\textbf{\SP.}\footnote{\href{https://huggingface.co/tceron/sentence-transformers-party-similarity-by-party}{https://huggingface.co/tceron/sentence-transformers-party-similarity-by-party}} This model based on \SV{} is fine-tuned such that representations of CMP quasi-sentences produced by the same party are close together, thus capturing information on the way parties express their claims, ideologies, and opinions \citep{ceron-etal-2022-optimizing}.
Like \SD{}, this model serves as a baseline trained on political data but not specifically on social media data.

\noindent
\textbf{\SH.} To fit the political social media domain of the dataset used in this study, we fine-tune \SV{} with tweets from the Twitter corpus following the procedure laid out in Section \ref{sec:hashtag-tuning}. To account for changes in the usage of Twitter, ensure evaluation on unseen data, and to simulate a realistic scenario, we fine-tune one model per evaluated election year on tweets from the four years leading up to the respective election year targeted in the evaluation.

For this, we sample all Tweets with hashtags that occur for at least three parties and at least 50 times across parties from the four years leading up to the respective election year. This results in 4,450 and 8,829 hashtags for 2017 and 2021 respectively. We randomly sample positive pairs from the cross-product of all tweets with the same hashtag, and negative samples from the cross-product of tweets with the respective hashtag and tweets with no co-occurring hashtag, ensuring balanced training sets. Given limited resources, we cap the total number of training examples per model at 2.5M examples.
We use contrastive loss \citep{contrastiveloss} as our training objective.

\subsubsection{Distance Aggregation Definitions}
To assess the effect of our hashtag-informed distance aggregation as described in Section \ref{sec:topic-aggregation}, we compare the performance of this distance aggregation approach to using averaged cosine distances between the sentence embeddings of all possible tweet pairs of two given parties as inter-party distances. Formally, we apply Equation \ref{eq:dist} on the whole sets of tweets of a given pair of parties instead of applying it per hashtag first and then averaging. Both settings use \SH{} and we report results for both election years. 

\subsection{Results and Discussion}

\begin{table}[tb!]
    \centering
\begin{tabular}{lll}
\hline
\textbf{Model} & \textbf{2017}   & \textbf{2021}   \\\hline
\SV{}    & -0.461 & 0.369  \\
\SD{}    & -0.025 & 0.352  \\
\SP{}    & \phantom{-}0.649  & 0.813* \\
\SH{}    & \phantom{-}0.803* & 0.817* \\\hline
\end{tabular}
    \caption{Experiment 1: Mantel correlation results using the full dataset available in the respective year. * $p<0.05$.}
    \label{tab:results_full}
\end{table}

Table \ref{tab:results_full} shows the correlations of the aggregated distances per party to the respective manifesto ground truth. 
For the evaluation of the distance matrices, we find that the only models that reach significantly high Mantel correlations are the party-tuned model \SP{} in 2021 with a Mantel correlation of 0.813, and our political social media domain-specific models, \SH{} in both years (0.803 for 2017, and 0.817 for 2021). Moreover, the \SH{} models are the only models yielding comparable results for both years. The other two representation models, \SV{} and \SD, in contrast, yield considerably lower correlations, failing to reach significance for both election years.

Overall, these results suggest that given the need 
to capture meaningful distances between parties on Twitter across years, representation models fine-tuned to leverage insights on how politicians use the platform to fit the political social media domain are preferable. The contrast between \SD{} and \SH{} in particular shows that while both models are trained on political information, the relationship between statements learned only from manifestos does not align well with the political social media domain.

As shown in Table \ref{tab:dist-res}, given genre-specific representations, our topic-informed distance aggregation method produces distances with a high and significant correlation to the manifesto distances, while simple averaging does not. We take this as evidence that to retrieve meaningful ideological distances between parties, it is essential i) to have representations that can capture genre-specific differences on a sentence/tweet level, and ii) to measure these differences in an informed way.

\begin{table}[]
    \centering
    \begin{tabular}{lll}
    \hline
       \textbf{Distance definition} & \textbf{2017} & \textbf{2021} \\\hline
       Average  & 0.347 & 0.354 \\
       Topic Aggregation  & 0.803* & 0.817*\\\hline
    \end{tabular}
    \caption{Experiment 1: Mantel correlations for \SH{} with our distance aggregation approach and using average distances. * $p<0.05$.}
    \label{tab:dist-res}
\end{table}

\section{Experiment 2: Random Subsampling}
Following the insights from our first experiment, we assess how stable the positioning retrieved using the \SH{} models in our proposed workflow are by randomly subsampling tweets.
In doing so, we simulate lower-resource scenarios of regional elections and political landscapes of countries with considerably fewer tweets available.
Our question is how low the number of Tweets can be before the performance degrades considerably.

\subsection{Experimental Setup}
For both evaluation years, we randomly sample decreasing percentages of the full dataset, starting from 87.5\% of the tweets being sampled down to 12.5\% in steps of 12.5\%.

As in the previous experiment, given the \SH{} embeddings of the tweets of all hashtags that occur across all parties, we retrieve inter-party distances and evaluate them against the respective manifesto distance matrix.   
To ensure representative results, per subsampling setting, we conduct five runs with different random seeds. 

\subsection{Results and Discussion}
\begin{figure}
    \centering
    \includegraphics[width=\linewidth]{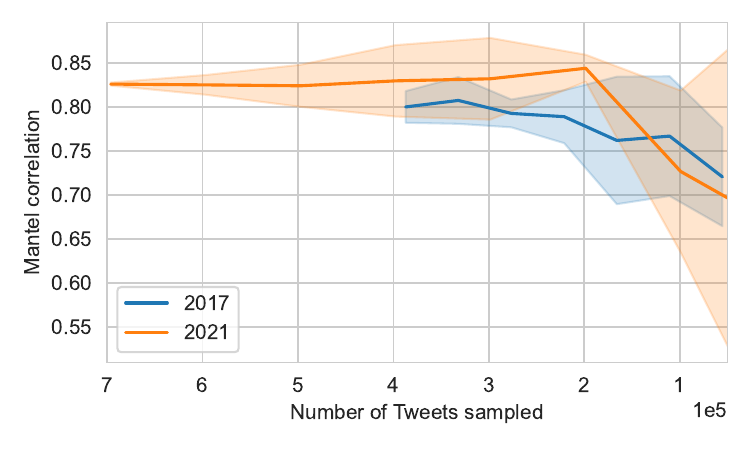}
    \caption{Experiment 2: Mantel correlation results for \SH{} for subsampled subsets of the dataset for the election years 2017 and 2021. Average over five runs with standard deviation.}
    \label{fig:random-pruning}
\end{figure}

The average results across the five runs with standard deviation for each subsampling step for both years are shown in Figure \ref{fig:random-pruning}. 
For both election years, we find fairly stable results for the higher numbers of tweets sampled, with Mantel correlations approximating 0.8 from about 200k tweets.
Unsurprisingly, there is a downward trend for these lower numbers of tweets as well as larger standard deviations across runs. Still, we find that our method still mostly produces distances significantly correlating with manifesto distances. Even in the  scenarios with the highest standard deviations (<100k tweets for the evaluation year 2021), we find that three out of the five runs produce significant Mantel correlations of $>0.7$.

These results indicate that our methodology remains generally robust in low(er) resource scenarios. The higher standard deviations, however, indicate that it may matter \textit{which} tweets are sampled. As our method assumes a topically diverse, representative sample of tweets (and hashtags) to retrieve meaningful aggregated party positioning, the main risk of drawing smaller samples is that they may not to be as topically diverse and representative as larger ones.

\section{Experiment 3: Temporal Sampling}
The discourse on Twitter can be reasonably expected to be more driven by short-term political topics and idiosyncratic events in some time periods than in others. As our method relies on a representative, broad topical distribution, we expect this to have an impact on its produced distances.

In this experiment, we thus assess how fine-grained the periods are for which we can make predictions using our method. 

\subsection{Experimental Setup}\label{sec:temporal_sampling}
We sample all tweets available in restricted time spans within election years in two settings: (a) We consider only the months leading up to the respective election month (September for German federal elections). We sample Tweets in increasingly shorter time spans from 9 months (Jan-Sep), over 6 months (Mar-Sep) to 3 months (Jun-Sep). (b) We sample tweets of individual months leading up to the election. We show the number of Tweets per month in Figure \ref{fig:tweets-monthly} in the Appendix.
For every setting in (a) and (b), we retrieve interparty distances given the \SH{} embeddings of the tweets of all hashtags that occur across all parties and evaluate them against the respective manifesto distance matrix.

\subsection{Results and Discussion}

\begin{table}[tb]
    \centering
    \begin{tabular}{lrrr}
    \hline
     & \multicolumn{1}{l}{\textbf{Jan-Sep}} & \multicolumn{1}{l}{\textbf{Mar-Sep}} & \multicolumn{1}{l}{\textbf{Jun-Sep}} \\\hline
2017 & 0.791*                      & 0.799*                     & 0.770*                      \\
2021 & 0.781*                     & 0.769*                     & 0.888*   \\\hline                 
\end{tabular}
    \caption{Experiment 3 (a): Mantel correlations for \SH{} for the temporal subsampling experiment sampling in time intervals leading up to the election month (September) for the election years 2017 and 2021. * $p<0.05$.}
    \label{tab:results-time}
\end{table}

The results for (a) are presented in Table \ref{tab:results-time}. While the Mantel correlation decreases for all restricted timespans except for Jun-Sep 2021 compared to  the respective full year, all correlations remain high and significant. Sampling Jun-Sep leads to the lowest correlations in 2017 and Mar-Sep to the lowest correlations in 2021. Interestingly, while there might be a proximity effect for 2021, we do not find that tweets become more aligned with manifestos closer to the elections in 2017.

\begin{figure}
    \centering
    \includegraphics[width=\linewidth]{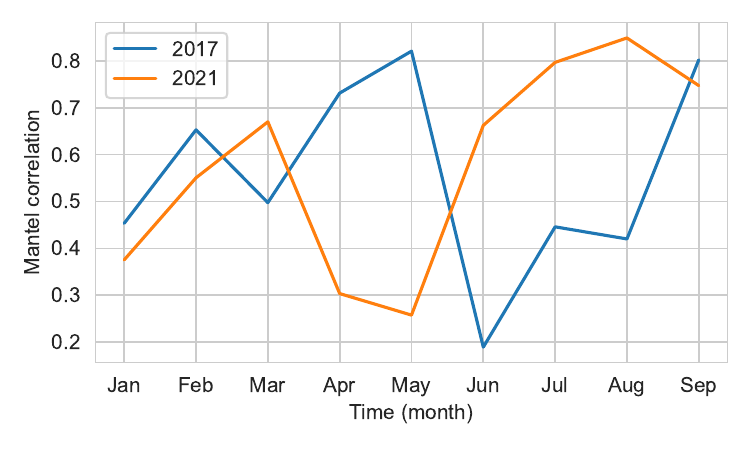}
    \caption{Experiment 3 (b): Results for \SH{} sampling tweets of individual months in the election years 2017 and 2021.}
    \label{fig:results-month}
\end{figure}

Figure \ref{fig:results-month} shows the results for (b), sampling tweets per month. For both years, the correlations to the manifestos vary considerably from month to month. While there is a trend of high correlations in the election month, there are periods between the beginning of the respective year and the election month. January, April, and May for 2021, and January, June, July, and August for 2017 have correlations lower than 0.5. These drops coincide with the lower correlations when sampling the respective months in (a). 

Overall, these results indicate that, on the one hand, given all tweets sampled from at least three months before the given election, positioning with reasonable correlations to the manifesto can be retrieved (cf. Table \ref{tab:results-time}). We interpret this as an indication of a representative topical distribution being present to retrieve overall distances between parties from.
In a similar vein, the drops in correlation in some months indicate the discourse differing from the general topical distribution, focussing more on short-term political issues and idiosyncratic events. Analysis of the most used hashtags per month reveals that in April 2021, for instance, two hashtags related to the passing of two bills, one related to the COVID-19 pandemic (\#Infektionsschutzgesetz) and a rental cap (\#Mietpreisbremse) 
exhibit a frequency increase of factor 8.5 and 2.7 over the long-term average, respectively.   
\#Infektionsschutzgesetz has a particularly pronounced increase, jumping from being the 70th-most frequent hashtag long-term to being the 4th-most frequent hashtag in April 2021.

\section{Experiment 4: Group-based Sampling}
In this experiment, we assess the impact of selecting different groups of politicians when sampling tweets. We do this to investigate the stability of positioning with respect to tweets of only a subset of politicians being sampled. At the same time, this exemplifies a use case for our method: comparing how well different groups are aligned with the manifesto (i.e., the main party line). 

\subsection{Experimental Setup}
We sample all tweets available in the election year 2021\footnote{We only conduct this experiment for 2021 as only 4/6 parties elected in 2017 previously held seats in the Bundestag.} from (a) politicians newly elected into the German parliament in the respective year, (b) incumbent politicians (i.e. politicians with a seat in the German parliament) that got re-elected, and (c) incumbent politicians that did not get re-elected.
We hypothesize that incumbent politicians represent the party consensus better than newly elected politicians, as the parties nominate their candidates and incumbent politicians running for re-election are \textit{established} within the party. 

For all groups, we retrieve party distances given the \SH{} embeddings of the tweets of all hashtags that occur across all parties and evaluate against the 2021 manifesto distance matrix.

\subsection{Results}

\begin{table}[tb!]
\centering
\resizebox{\linewidth}{!}{%}
\begin{tabular}{lllll}
\hline
       & \textbf{Old}   & \textbf{Continuing} & \textbf{New}    & \textbf{All}    \\
       \hline
Mantel & 0.328 & \multicolumn{1}{r}{0.854*}     & \multicolumn{1}{r}{0.792*} & 0.817* \\
\hline
\end{tabular}}
\caption{Experiment 4: Mantel correlations for \SH{} for the election year 2021 using the tweets of not re-elected (Old), re-elected (Continuing), and newly elected (New) politicians. * $p<0.05$.}
\label{tab:results-elected}
\end{table}

The results are presented in Table \ref{tab:results-elected}. For both elected groups (\textit{Continuing} and \textit{New}), we find significant correlations to the manifesto distances. \textit{Continuing} reaches the highest correlation with 0.854, higher than the setting sampling all politicians (0.817), while \textit{New} shows a lower correlation (0.792) than the full setting. Only the setting using the tweets of not-re-elected (\textit{Old}) politicians fails to reach significant correlations.

These results indicate that re-elected incumbent politicians indeed reflect the main party line more than newly elected politicians or all politicians affiliated with a given party.
While the distances retrieved from incumbent not-re-elected (\textit{Old}) have a low correlation to the manifesto distances, this result does not suffice for the conclusion that their positions differ considerably from the party line, as \textit{Old} comprises much fewer tweets (roughly 45k vs. over 100k for \textit{New} and over 200k for \textit{Continuing}). As we can see high correlations even with comparable numbers of tweets (cf. Experiment 2), this effect requires further investigation. 

\section{Analysis}
\begin{figure}
    \centering
    \includegraphics[width=\linewidth]{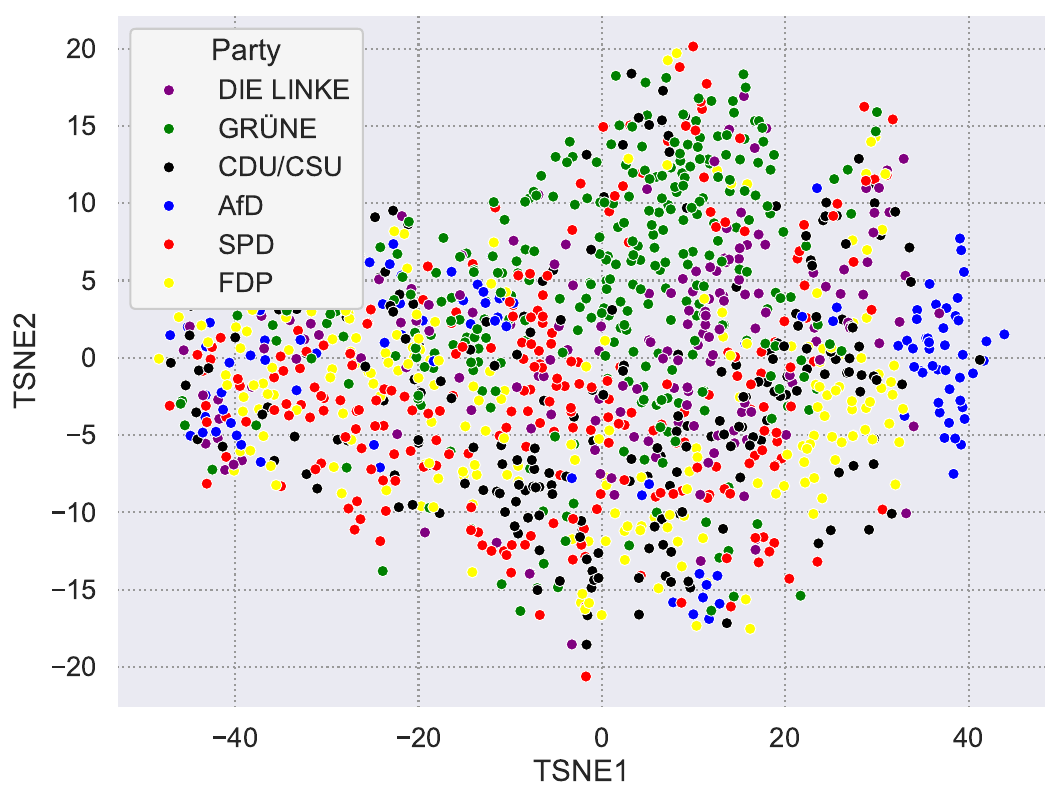}
    \caption{Two-dimensional projection of politician's \SH{} embedding centroids for the year 2021. Appendix \ref{appendix:party_names} introduces party acronyms and names.}
    \label{fig:tsne-decomposition}
\end{figure}

In our experiments, our focus has been at an aggregate level, predicting distances between parties. This raises the question of what our fine-tuned models capture at a lower level. 
To validate that our genre-tuned representations reflect ideological differences between members of different parties, we analyze the centroids of individual politicians' tweets from 2021 by projecting them into two dimensions using t-SNE \citep{tsne}.

As shown in Figure \ref{fig:tsne-decomposition}, there are regions in the projection where politicians of specific parties tend to cluster. Moreover, parties that ideologically are closer together at least in tendency are in each other's vicinity. For instance, the Greens' (GRÜNE) and the Left's (DIE LINKE), as well as AfD's and FDP's most predominant clusters are close to each other, respectively (cf. Figure \ref{fig:cmp-matrix}).
While there is no clear clustering separating the parties, we argue that this is expected, as (a) different politicians may have different foci and tweeting behaviors, (b) there is expected overlap between politicians of different parties, either by a tendency to talk about the same topics or by actual overlap in their expressed views.

\section{Conclusion}

In this paper, we propose and robustly evaluate a method to create representations for politicians' tweets which, relying on hashtag as a fine-tuning signal, proved successful in matching politicians' statements to the official lines of the respective parties, as encoded in the electoral manifestos. Furthermore, we have established the robustness of this method to low-resource scenarios via random sub-sampling, as well as with the sampling of time slices and groups of politicians. Such robustness is of particular relevance for interdisciplinary applications -- given that low-resource scenarios are a typical situation in the social sciences, making robustness a crucial desideratum. 

This workflow facilitates, for instance, the immediate tracking of shifts in positioning among politicians and parties following data collection. Moreover, given that it is increasingly important to follow how much politicians align or diverge their parties because it helps in holding them accountable or monitoring how extreme their positioning can become, this work is one step forward towards automatically analyzing political discourse on social media. 

A natural avenue for future research is to add interpretability to the approach, which would permit us to understand qualitatively where politicians align or diverge the most. 

Finally, we note that the models we propose in this paper assume that the social media messages use hashtags. While the role of hashtags in political communication on Twitter is well-established both theoretically and empirically, this begs the question of \textit{what to do when there are no hashtags available}, either when the authors decide against using them or when we want to move to a social media platform where hashtags are not in general use. Of course, it is possible to carry out distance aggregation with a distance measure that does not depend on hashtags. Appendix \ref{appendix:bonus-exp} presents a pilot study on such a measure, which indicates that such an approach is in principle possible, but that ignoring the strong signal from hashtags comes at a substantial cost in performance. 
More research is necessary to define an improved distance measure.

\section*{Limitations}
The main shortcoming of this work is the limited scope of the data and experiments. Firstly, our work only evaluates the method in one language for one political landscape (Germany/German), and for two election years.
In principle, the method of fine-tuning sentence encoder models based on and aggregating distances according to hashtag information can be applied to other languages and countries. The usage of individual hashtags and the overall usage of social media platforms by politicians, however, can be expected to differ considerably between countries and different periods, as the focus of the short-term political discourse can differ drastically.
Secondly, our method is based on politician's usage of only one social media platform, Twitter/X. While other platforms such as Facebook, Instagram, or TikTok may have a hashtag equivalent, it is not clear whether politicians use them comparably and thus whether our proposed method can be applied to other platforms.

Moreover, our work only considers the comparison to the political positioning extracted from manifestos as coordinates to assess the political positioning on social media, while there are other options such as voting advice applications or expert surveys. However, given the relevance of CMP-based measures within the (computational) political science community and, more importantly, the high agreement of ideological measures based on them with these other ground truth options \citep{silva-etal-2023-vaas}, we are confident that our work contributes to the assessment of the ideological overlaps and differences between a number of political text genres.

\section*{Ethical Considerations}
As the method presented in this work is designed to conduct large-scale investigations of how politicians ideologically position themselves on Twitter and relies on insights very specific to politicians and their use of social media, we do not believe this work to have considerable risks. While it is theoretically possible to also aggregate positions from other types of Twitter users, it does not allow for accurate profiling of individual users in the form presented here. We thus deem the chance of misuse of our method and models to be rather slim. 

\section*{Acknowledgements}

We acknowledge funding by the Deutsche Forschungsgemeinschaft (DFG) for the
project MARDY 2 (375875969) within the priority program RATIO.

\bibliography{anthology,custom}
% \bibliography{custom}

\appendix
\onecolumn
\section*{Appendix}
\section{Twitter corpus: details}

\begin{table}[h]
    \centering
    \begin{tabular}{lrr}
    \hline
    \textbf{Year} & \multicolumn{1}{l}{\textbf{Tweets}} & \multicolumn{1}{l}{\textbf{Politicians}} \\ \hline
     2013 & 573,953 & 668 \\
     2014 & 554,566 & 732 \\
     2015 & 601,370 & 838 \\
     2016 & 645,272 & 962 \\
     2017 & 870,007 & 1,187 \\
     2018 & 954,681 & 1,240 \\
     2019 & 1,019,515 & 1,321 \\
     2020 & 1,246,018 & 1,439 \\
     2021 & 1,423,427 & 1,587 \\\hline
     Overall & 10,220,767 & 1,924 \\\hline 
    \end{tabular}
    \caption{Dataset Statistics}
    \label{tab:dataset}
\end{table}

\section{German Party Acronyms \& Names}
\label{appendix:party_names}
\begin{table}[h]
    \centering
    \resizebox{\linewidth}{!}{%}
    \begin{tabular}{lll}
        \toprule
        \textbf{Acronym} & \textbf{Name} & \textbf{Translation} \\
        \midrule
        AfD & Alternative für Deutschland & Alternative for Germany \\
        CDU/CSU & Christlich-Demokratische Union/ & Christian Democratic Union/\\&Christlich-Soziale Union&Christian Social Union\\
        FDP & Freie Demokratische Partei & Free Demokratic Party\\
        GRÜNE & Bündis 90/Die Grünen & Alliance 90/The Greens\\
        LINKE & Die Linke & The Left\\
        SPD & Sozialdemokratische Partei Deutschlands & Social Democratic Party of Germany\\\bottomrule
    \end{tabular}}
    \caption{German Party Acronyms, Names and Translation of the Names}
    \label{tab:party_names}
\end{table}

\section{CMP Categories}
\begin{table}[h]
\centering
\resizebox{\textwidth}{!}{%
\begin{tabular}{llllll}
\hline
\textbf{Domain}  &  & \textbf{External Relations}          & \textbf{Freedom and Democracy} & \textbf{Political System} & \textbf{Economy}  \\\hline
Example Category &  & Internationalism: Positive           & Freedom and Human Rights       & Decentralization          & Market Regulation \\ \hline
\end{tabular}}

\bigskip
\resizebox{0.9\textwidth}{!}{%
\begin{tabular}{lllll}
\hline
\textbf{Domain}  &  & \textbf{Welfare and Quality of Life} & \textbf{Fabric of Society}     & \textbf{Social Groups}                   \\\hline
Example Category &  & Environmental Protection             & Traditional Morality: Negative & Labour Groups: Positive                   \\ \hline
\end{tabular}}
\caption{The seven CMP policy domains \citep{manifesto} and example categories per domain.}
\label{tab:categories}
\end{table}

\newpage
\section{CMP Distance Matrix}
\begin{figure}[h]
    \centering
    \includegraphics[width=0.7\linewidth]{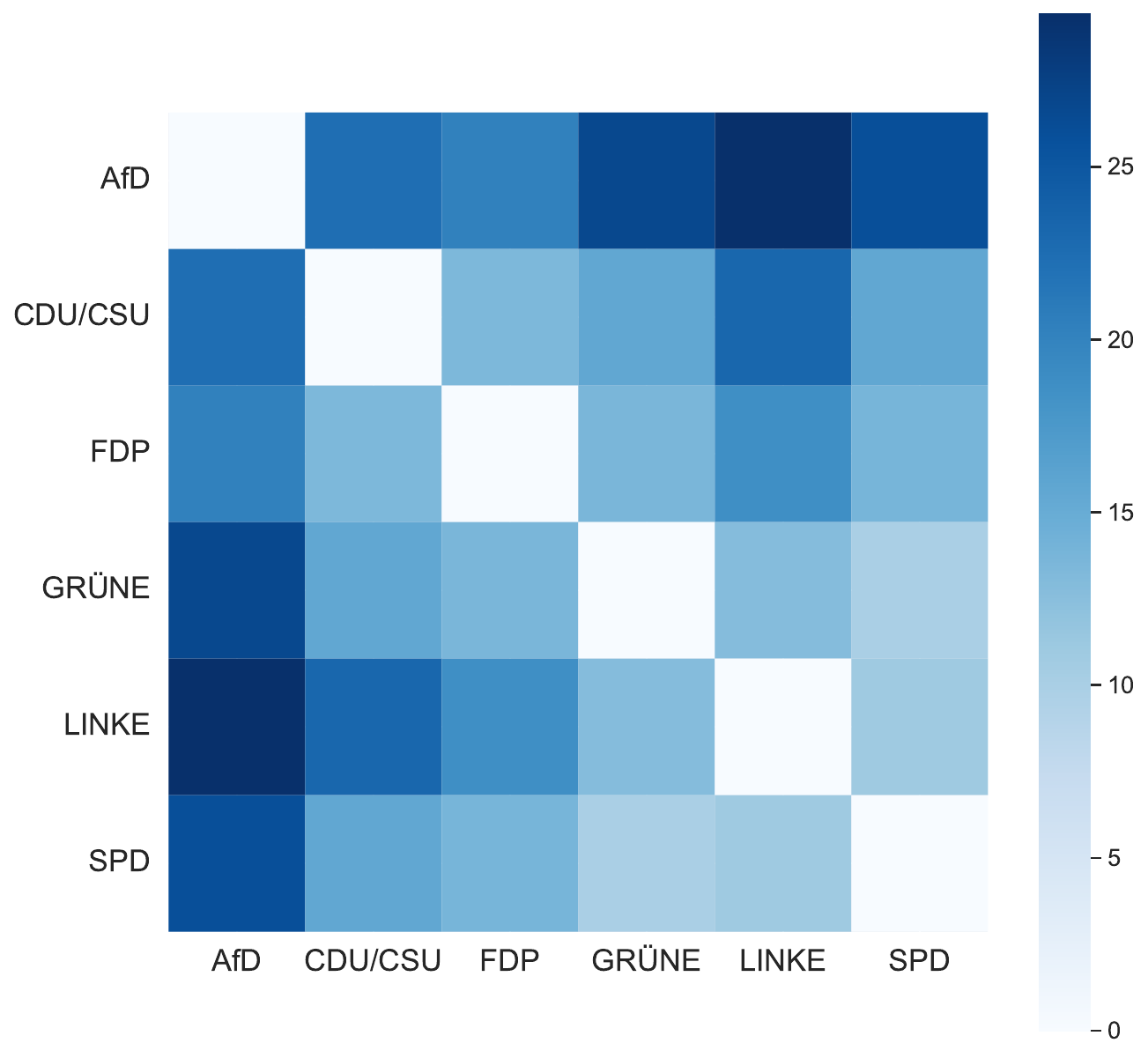}
    \caption{CMP distance matrix of the election year 2021. The similarity between the parties increases with lighter colors.}
    \label{fig:cmp-matrix}
\end{figure}

\section{CMP annotation example}
\label{appendix:cmp-annot}
\begin{figure}[h]
    
    \small\textbf{DE} \colorbox{yellow}{Daraus erwächst Verantwortung.} \colorbox{yellow}{In Europa,} \colorbox{cyan}{aber auch für die Schwächeren in unserer Gesellschaft}\\
    \textbf{EN} \colorbox{yellow}{This gives rise to responsibility.  } \colorbox{yellow}{In Europe,} \colorbox{cyan}{but also for the weaker members of our society.}
    \caption{Annotation example for three quasi-sentences of the 2013 SPD party manifesto with translations. Example from the manifesto corpus \citep{manifestocorpus}.\\\\Legend: \colorbox{yellow}{Domain 6: Fabric of Society} \colorbox{cyan}{Domain 5: Welfare and Quality of Life}}
    \normalfont{}
    \label{fig:cmp-annot}
\end{figure}

\section{Figure 1: Sources}
\label{appendix:fig1}
\begin{itemize}
    \item Manifesto AfD: \url{https://www.afd.de/wp-content/uploads/2017/04/2017-04-12_afd-grundsatzprogramm-englisch_web.pdf}, p. 62
    \item Manifesto Die Linke: \url{https://www.die-linke.de/fileadmin/download/wahlen2017/wahlprogramm2017/die_linke_wahlprogramm_2017.pdf}, p.64
    \item  Tweet IDs, from top: 816234178601684992, 815812840586170368, 929669873852088320, 816293066059476992
\end{itemize}
All texts (except the AfD manifesto) originally in German, our own translation into English.

\newpage
\section{Experiment 3}

\begin{figure}[h]
    \centering
    \includegraphics[width=0.6\columnwidth]{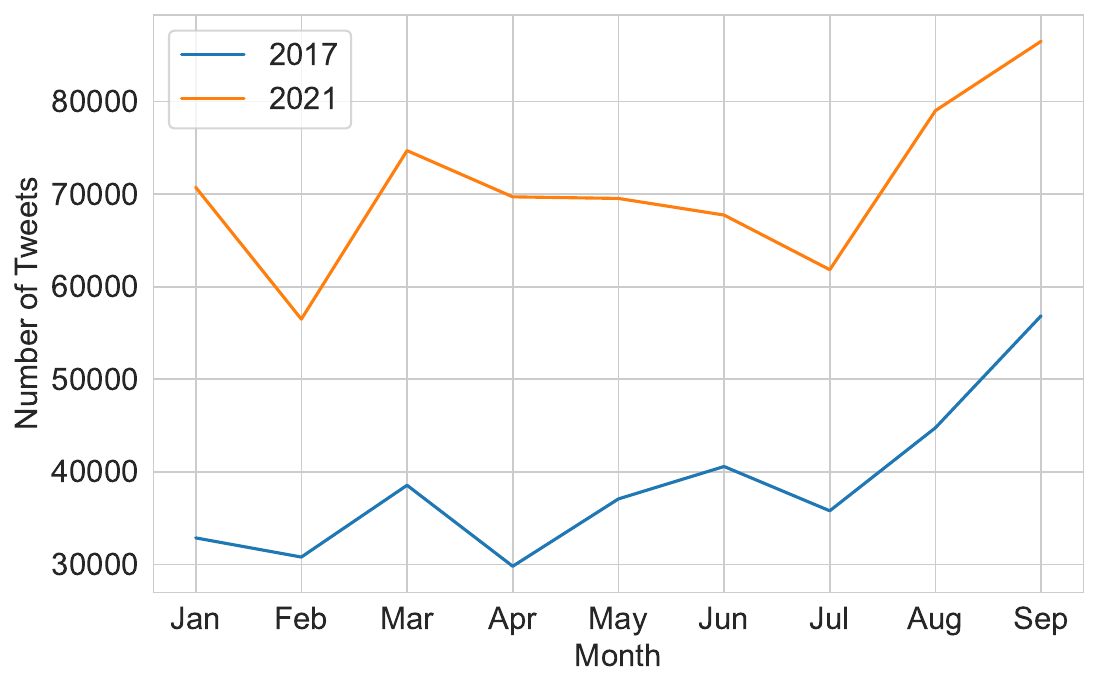}
    \caption{Experiment 3: Number of tweets per month per election year.}
    \label{fig:tweets-monthly}
\end{figure}

\section{Bonus Experiment: Alternative Distance Definition}
\label{appendix:bonus-exp}
In this additional experiment, we assess the usefulness of our fine-tuned models in situations where there may be no hashtags available in the texts. Since the distance measure introduced in Section \ref{sec:topic-aggregation} relies on hashtags and simple averaging of all cosine distances does not yield meaningful results (cf. Experiment 1: Table \ref{tab:dist-res}), this forces us to use another informed distance definition. We thus adapt an interparty similarity definition proposed in earlier work \citep{ceron-etal-2022-optimizing} to a quasi-metric. Since this definition leverages inter- and intra-party nearest neighbors per sentence, it does not rely on hashtags.

\subsection{Experimental Setup}
\textbf{Data.} We split the data for the election years 2017 and 2021 used throughout this paper into three subsets: (1) Only tweets that do not contain hashtags. (2) Only tweets that contain hashtags. This is equivalent to the subsets of the data used throughout this paper. (3) All tweets.

\noindent
\textbf{Models.}
We use \SH{} and \SP{}, as described in Section \ref{sec:hashtag-tuning}.

\noindent
\textbf{Distance Definition.}
We adapt \citet{ceron-etal-2022-optimizing}'s proposed similarity measure they dub \textit{twin matching}. To derive the similarity between two parties, per sentence, twin matching aggregates the similarity of its most similar sentence (in our case tweet) across parties (\textit{inter-party twin}), normalized by the similarity of the most similar sentences within the parties (\textit{intra-party twin}) respectively, given the sets of sentences of the two parties.
The underlying intuition is that, given a fitting sentence embedding model, the sentence and its twins should be thematically similar in most cases. 

Formally, given a sentence $s$ in a collection of sentences $T$, in our case tweets, the neighbor or twin of a sentence, $tw(s,T)$ is defined as 
\[
    tw(s,T)=\arg\max_{t\in T}\cos (s,t)
\]
where $\cos(s,t)$ is the cosine similarity of $s$ and $t$.
The maximum inter-collection similarity $C(P)$ of a set of tweets of a party $P$ is
\[
    C(P) = \max_{p,p' \in P \And p\neq p'} \cos(p,p')\text{.}
\]
\noindent
The inter-party similarity between the sets of tweets of parties the parties, $P_1$ and $P_2$: 
\[
    sim(P_1,P_2)=\sum_{s\in P_1} \frac{\cos(s, tw(s, P_2))}{|P_1|(C(P_1) + C(P_2))}
\]

Since $sim(P_1,P_2)\neq sim(P_2,P_1)$ unless $P_1=P_2$ and $sim(P_1, P_1)\neq 1$ due to $C(P)$, we adapt \citet{ceron-etal-2022-optimizing}'s similarity measure to a quasi-metric:

\[
    dist(P_1, P_2)=
\begin{cases}
    0, & \text{if } P_1=P_2\\
    1-\frac{sim(P_1,P_2) + sim(P_2,P_1)}{2}, & \text{if } P_1 \neq P_2
\end{cases}
\]

\noindent
\textbf{Evaluation.} 
We follow the same evaluation procedure as in the other experiments in the paper: Per model and dataset split, we use the Pearson-normalized Mantel test to evaluate the resulting distances against CMP distances of the respective election year.

\subsection{Results}

Table \ref{tab:bonus-res} shows the Mantel correlations for the three splits per election year. For both election years and across data subsets, we find that our genre-specific models, \SH{} yield higher and more consistent correlations than \SP{}. \SP{} fails to reach significance in all settings of both election years with low correlations (0.081--0.362). \SH{}, in contrast, yields high significant correlations (0.742--0.927) in all settings except for the subset containing only tweets without hashtags in 2017. For that subset in 2021, however, we find a high significant correlation (0.742). 
The correlations for subsets containing (only) tweets with hashtags (\textit{Hashtags} and \textit{Full}) are systematically higher than for the subsets where only tweets without hashtags are available.

Overall, we take this as an indication that, given a fitting distance definition, our genre-specific models can be useful even if no hashtags are available. At least with the specific distance definition present in this experiment, however, the results in these cases appear to be sensitive to the number of data points used, as the difference between 2017 and 2021 shows.

\begin{table}[!h]
\centering
\begin{tabular}{lllllll}
\toprule
       & \multicolumn{3}{c}{2017}        & \multicolumn{3}{c}{2021}        \\
\cmidrule(lr){2-4}\cmidrule(lr){5-7}
Subset & No Hashtags & Hashtags & Full   & No Hashtags & Hashtags & Full   \\
\midrule
\SH{}     & 0.576       & 0.831*   & 0.903* & 0.742*      &  0.926*  & 0.927* \\
\SP{}     & 0.081       &  0.250   & 0.088  & 0.240       & 0.362  & 0.294 \\\bottomrule
\end{tabular}
\caption{Mantel correlations for \SH{} and \SP{} with twin matching distance for subsets of the data containing only tweets (a) without hashtags (\textit{No Hashtags}), (b) with hashtags (\textit{Hashtags}), and (c) the full dataset (\textit{Full}) for the election years 2017 and 2021.}
\label{tab:bonus-res}
\end{table}

\section{Model Parameters \& Training Details}

For all models other than \SH{}, we use the implementation available through the Hugging Face Hub\footnote{\href{https://huggingface.co/}{https://huggingface.co/}} and the transformers\footnote{\href{https://huggingface.co/docs/transformers/en/index}{https://huggingface.co/docs/transformers/en/index}} library.

\noindent
\textbf{\SH{}.}
Both models (2017 and 2021) were fine-tuned on 2.5M examples using the following hyperparameters:
\begin{itemize}
    \item Pre-trained SBERT model: \texttt{paraphrase-multilingual-mpnet-base-v2}\footnote{For model parameters and additional details, consult \href{https://huggingface.co/sentence-transformers/paraphrase-multilingual-mpnet-base-v2}{https://huggingface.co/sentence-transformers/paraphrase-multilingual-mpnet-base-v2}.}
    \item 2 Epochs
    \item Batch Size 32
    \item Warmup Steps 1000
    \item AdamW optimizer
    \item Learning Rate $2\mathrm{e}{-5}$
    \item Loss Function: Contrastive Loss
    \item sentence-transformers\footnote{\href{https://www.sbert.net/}{https://www.sbert.net/}} implementation
\end{itemize}
Per model, the training took approx 6 hours on an NVIDIA RTX A6000. We save the best model w.r.t. the validation loss.
Our fine-tuned models are publicly available through the Hugging Face Hub\footnote{\href{https://huggingface.co/mmmaurer/sbert-hashtag-german-politicians-2021}{https://huggingface.co/mmmaurer/sbert-hashtag-german-politicians-2021} and \href{https://huggingface.co/mmmaurer/sbert-hashtag-german-politicians-2017}{https://huggingface.co/mmmaurer/sbert-hashtag-german-politicians-2017}}.

\section{Implementation Details}

We use the following libraries for data processing, experiments, and evaluation:
\begin{itemize}
    \item Training
    \begin{itemize}
        \item PyTorch\footnote{\href{https://pytorch.org/}{https://pytorch.org/}}
        \item sentence-transormers
    \end{itemize}
    \item Evaluation: mantel\footnote{\href{https://github.com/jwcarr/mantel}{https://github.com/jwcarr/mantel}}
    \item Data processing: pandas \footnote{\href{https://pandas.pydata.org/}{https://pandas.pydata.org/}}, polars \footnote{\href{https://pola.rs/}{https://pola.rs/}}
\end{itemize}

For the full details and our code, see our GitHub repo: \href{https://github.com/mmmaurer/toeing-the-party-line}{https://github.com/mmmaurer/toeing-the-party-line}.

\end{document}